# Bayesian Learning of Clique Tree Structure

**Cetin Savkli, J. Ryan Carr, Philip Graff, Lauren Kennell**
The Johns Hopkins Applied Physics Laboratory, Laurel, MD, USA

**Abstract -** *The problem of categorical data analysis in high dimensions is considered. A discussion of the fundamental difficulties of probability modeling is provided, and a solution to the derivation of high dimensional probability distributions based on Bayesian learning of clique tree decomposition is presented. The main contributions of this paper are an automated determination of the optimal clique tree structure for probability modeling, the resulting derived probability distribution, and a corresponding unified approach to clustering and anomaly detection based on the probability distribution.*

**Keywords:** Categorical, Probability, Anomaly, Clique Tree, Clustering

## 1   Introduction

With the rapid growth of categorical data available for analysis, the need for robust statistical approaches is becoming ever more critical. Unlike numerical data (such as weather or astronomical data), much of the data found in social networks, and the web in general, is categorical in nature. While methods for analysis of numerical data are well established, methods used for analysis of categorical data are more varied and still developing.

One of the challenges in the analysis of categorical data is a lack of a natural distance metric that most statistical learning algorithms rely on. This problem has led to numerous proposals and a comparative analysis of alternatives can be found in [1]. While the lack of a natural distance metric is a problem, it is also known that as the dimensionality of the attribute space increases the distance metrics become less and less useful, a fact that is also known as the "curse of dimensionality." In particular, the curse of dimensionality implies that in high dimensions the data becomes sparse (thinly spread over the total event space) and thus most of the data becomes equally anomalous. Categorical data, such as cyber or financial transactions, is often high dimensional and can easily comprise dozens of attributes. Therefore, identifying anomalies becomes a challenging task in many categorical data sets. Rule based or ground truth based classification approaches can be used to detect predefined event classes, but these methods have limited utility to detect new anomalies, meaning the rare events which have not been previously characterized. However, the anomalies of greatest interest may be new anomalies, and in fact they may be one-time events which do not form a cohesive class on which to train a classifier.

The inability to reliably identify anomalies has practical consequences as human inspection of large data sets for anomalies is a time-consuming activity and impractical on large data sets. Therefore, it is desirable to develop robust analytic approaches that do not require ground truth, do not rely on a distance metric, and that can handle the high dimensionality of the categorical data.

In this work, we explore a probabilistic approach to data representation that addresses the challenges described above. The approach is based on constructing a joint probability distribution using a structure called a clique tree (also known as a junction tree). The clique tree expresses dependencies in a high dimensional attribute space and can be used to make probabilistic inferences about data ([2], [3]). The clique tree approach is analogous to density estimation for numerical domains, but is more general as it can be used to infer probabilities of both numerical and categorical data.

Clique tree structures are an active area of research. Research about clique trees in literature has generally focused on inference algorithms as well as building of clique trees with a narrow width for fast inference. For instance, searching for a clique tree that satisfies various compactness criteria is discussed in [4], and clique tree based inference is discussed in [5] and [6].

The focus in this paper is the determination of an optimal clique tree structure which best represents attribute dependencies within the data and thus the information content of the data. The main contributions of this paper are:

- An automated and parameter-free determination of an optimal clique tree structure for probability modeling using Bayesian learning from data;

- A unified approach to clustering and anomaly detection based on the derived probability distribution.

In the following sections, we present the construction of a probability distribution, specification of the optimal clique tree structure directly from data, and applications to anomaly detection and clustering. Using a publicly available categorical data set, we present results which show that Bayesian learning can be used to construct an optimal clique tree that maximizes the probability of observed data while providing inference capability for unseen data.

## 2 The joint probability distribution

In high dimensional data sets, there is generally insufficient data from which to characterize probabilities; the available data points are spread too thinly over a very large space of possible attribute combinations. However, if it is known that some subsets of the variables are independent from other subsets of variables, a joint probability distribution in a high dimensional space can be decomposed into a product of lower dimensional probabilities ([2], [3]). Within low dimensional spaces, the data is more concentrated and a probability distribution can be successfully derived. Therefore, as the first step in deriving a joint probability distribution, dependencies will be characterized using the mutual information. The choice of mutual information as a metric for correlation is motivated by the generality of this metric in handling categorical and numerical data. However, it is possible to use another correlation metric as the approach is independent of a chosen metric. From the mutual information, an optimal structure for clique trees is obtained, and the probability model is based on this optimal clique tree structure.

### 2.1 Mutual information graph

For two variables $X$ and $Y$, the mutual information $I(X,Y)$ is defined by:

$$I(X,Y) = \sum_{x \in X} \sum_{y \in Y} p(x,y) \log\left(\frac{p(x,y)}{p(x)p(y)}\right) \quad (1)$$

A normalized version of the mutual information [7] can be used to establish the degree of dependence between attributes. The normalized mutual information (NMI) is given by

$$M(X,Y) = \frac{I(X,Y)}{\min(H(X),H(Y))} \quad (2)$$

where variable entropy $H$ is defined as

$$H(X) = -\sum_{x \in X} p(x) \log(p(x)) \quad (3)$$

The NMI varies between 0 and 1, where 0 indicates independence while 1 implies complete dependence.

To illustrate the ideas in this paper, a data set of mushrooms will be used [8]. The data set contains 8,124 mushrooms, each characterized by 22 attributes including color, shape, odor, etc., which are denoted in this paper by "a1" through "a22". For each pair of attributes, the NMI is computed, and the distribution is shown in Fig. 1. The distribution indicates that there are a small group of attributes with strong dependence while most attribute pairs have weaker dependence.

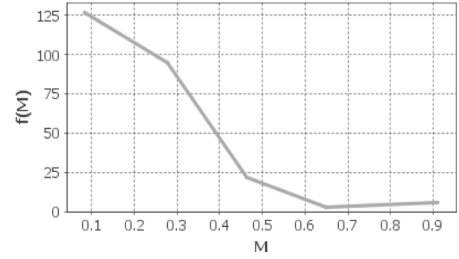

Fig. 1. Number of attribute pairs (f(M)) with a given mutual information (M) in the mushroom data set.

The mutual information results can be represented as a weighted graph. The 22 attributes are the graph nodes, and the link between each pair of nodes is weighted with the NMI of that attribute pair. Links can then be pruned with a threshold, so that only the links that indicate strong dependency are retained. Specification of this threshold is a key focus of this paper and is discussed in detail in Section 2.3.

### 2.2 Clique tree and joint probability model

For ease of illustration, we continue the derivation of a clique tree and probability model using a hypothetical data set with only six attributes $\{a, b, c, d, e, f\}$. The original NMI graph on these attributes would contain 15 weighted links. Suppose a threshold is chosen so that only the links in Fig. 2 are retained.

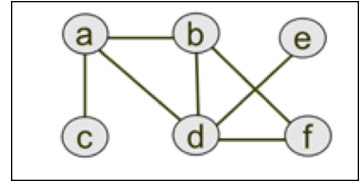

Fig. 2. Example of a pruned mutual information graph.

The pruned mutual information graph provides a basis for constructing a clique graph and clique tree. The first step is to find chordless cycles in the pruned mutual information graph and fix them. This is a necessary condition for the clique tree to satisfy the running intersection property, which guarantees that the clique tree will provide a joint probability distribution that is normalized [9]. A chordless cycle is a cycle such that nodes on the periphery have no direct connection to each other except for the nodes which are adjacent in the cycle. The pentagon-shaped cycle in Fig. 3 is an example of a chordless cycle. Fixing the chordless cycle can be accomplished by introducing the links shown on the right side of the figure.

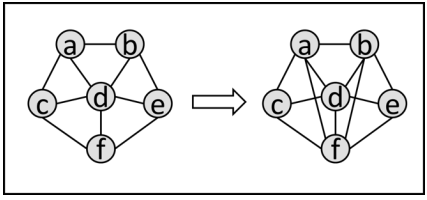

Fig. 3. On the left, the pentagon outline is a chordless cycle. This is fixed by adding the two additional edges shown at right.

After chordless cycles are repaired in the mutual information graph, we construct the clique graph. To form a clique graph, the maximal cliques of the input graph become the nodes of the clique graph. For instance, in the graph in Fig. 2, the node set $\{a,b,d\}$ is a maximal clique, and therefore $\{a,b,d\}$ becomes a node in the clique graph. Two clique graph nodes are linked if the cliques have at least one underlying node in common. We also label the link by those overlapping nodes, which are called the separator set. Lastly, to construct the clique tree from the clique graph, we use the minimum spanning tree algorithm where the link distances are measured in inverse of the separator set size [9]. The right side of Fig. 4 is the clique tree for the graph from Fig. 2. The ovals represent the maximal cliques, and the rectangles are the separator sets on the links.

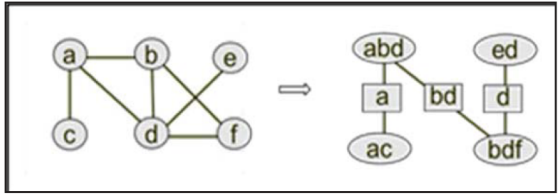

Fig. 4. The mutual information graph from Fig. 2 and its clique tree representation.

The derived probability model is easily read from the clique tree. For any data point $(a,b,c,d,e,f)$, its probability is given by the product of the probabilities within subspaces corresponding to maximal cliques, divided by the probabilities of the subspaces given by the separator sets:

$$P(a,b,c,d,e,f) = \frac{P(a,c)P(a,b,d)P(b,d,f)P(e,d)}{P(a)P(b,d)P(d)} \qquad (4)$$

Some of the benefits of this probability decomposition are immediately clear from its structure. First, the right side of equation (4) contains only lower dimensionality probabilities. The lower dimensional probabilities can be more reasonably inferred from available data, which addresses the problem of data sparsity and the curse of dimensionality. Second, the subspaces in the numerator of equation (4) consist of highly dependent attributes. When attributes are dependent, unexpected combinations of data values will immediately stand out; they will violate the expected behavior encoded in the attribute dependency. For instance, an unusual combination of values for $(b,d,f)$ would stand out as an anomaly in that subspace. The factor $P(b,d,f)$ will have a relatively low value, and in turn this will tend to cause $P(a,b,c,d,e,f)$ to have a low value as well. Therefore, the probability model correctly finds anomalies because it judges a data point by its conformance to expected variable dependencies. We will return to this idea in Section 3. In Section 2.3, we focus on determining the optimal mutual information threshold, and its relationship to the generalization capability of the distribution.

## 2.3 Generalization capability of the distribution

An immediate question regarding the joint probability distribution is related to generalization capability. The problem of generalization is a familiar one from other areas of data analysis. For example, fitting data to a curve presents a trade-off between the accuracy of the fit and how well it explains the data points not yet measured. In the case of clique tree decomposition, the problem is related to the extent of the pruning of the mutual information graph that forms the basis of the clique tree. Consider the two extreme cases:

Case 1: If the NMI threshold is set to 0, all of the mutual information links are retained. The resulting probability distribution is the full joint distribution. The probability of any data point becomes equal to how frequently that data point is observed, and any data point not previously seen will be considered to have probability of zero. The left figure in Fig. 5 shows this type of distribution. In short, this threshold is an overfitting of the data. This is again why high dimensionality and data sparsity is clearly problematic for modeling a probability distribution, since there are presumably many combinations of attributes which are not inherently anomalous but are assigned a zero probability.

Case 2: If the NMI threshold is set to 1, all of the mutual information links are removed. This limit corresponds to fitting the data assuming that all dimensions are independent (also known as Naïve Bayes assumption). Treating attributes as independent leads to a distribution that covers the entire parameter space and corresponds to an overly smooth distribution. According to this distribution, shown on the right side of Fig. 5, all data has small and non-zero probability.

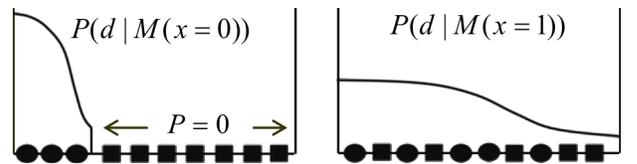

Fig. 5. Left: Setting mutual information threshold $x = 0$ results in unobserved data (squares) having zero probability. The average probability (P) of observed data (circles) is maximized, and entropy (S) is minimized. Right: Setting $x = 1$ results in all data points (observed and unobserved) being assigned positive probability, which maximizes the entropy.

Neither of the extreme cases is satisfactory. A threshold is needed which balances the competing goals of fidelity to observed data, and allowance for previously unseen data. In order to choose an optimal threshold $x$, we seek to maximize $P(M(x)|D)$ where $D$ is the set of observed data and $M(x)$ is the probability model derived from the clique tree corresponding to threshold $x$. While each threshold leads to a specific NMI graph, the relationship is not reversible. Threshold $x$ lives in a continuum while there are a finite set of trees that can be constructed using these thresholds. Using Bayes' Theorem, the posterior distribution for the model $M(x)$ specified by an NMI threshold $x$ can be expressed as

$$P(M(x)|D) = \frac{P(D|M(x))P(M(x))}{P(D)} \quad (5)$$

The prior $P(M(x))$ can be assumed to be uniform, and $P(D)$ is a normalization factor independent of $x$. Therefore, the problem is equivalent to maximizing $P(D|M(x))$, which is the product of individual data probabilities. However, even though the probability of the observed data needs to be maximized, using all available data for this purpose leads to over-fitting of the distribution leading to an optimal threshold of $x = 0$. It is necessary to force the distribution to assign mass to a more expansive set of data points than those it trains on, but not across the entire attribute space.

The solution is to divide the observed data into a training set and a test set. With this partition, the quantity we wish to maximize is given by

$$P(D|M(x)) = P(D_{Train}|M(x))P(D_{Test}|M(x)) \quad (6)$$

If any of the test data is assigned zero probability, the right side of equation (6) becomes zero, and the corresponding $x$ will therefore not be chosen as optimal. The idea is shown pictorially on the left side of Fig. 6.

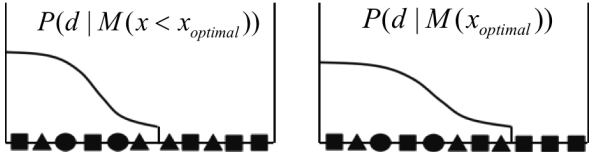

Fig. 6. Circles represent training data, triangles represent test data, and squares represent unobserved data. Left diagram: If the threshold is too low, some of the test data will have probability of zero. This threshold is rejected by maximizing $P(D|M(x))$ in equation (7). Right diagram: The threshold is just large enough to assign positive probability to all of the training and test data.

When the NMI threshold is low, the distribution will accommodate the training data, but it will be too compact to explain the test data. At the optimal threshold, shown in the right side of Fig. 6, the distribution assigns positive probability to the training data and the test data, and probably also to some of the possible data points which are still unseen. If the threshold were higher than optimal, $P(D|M(x))$ would start to decrease, since more of the mass would be assigned to the unobserved data. In summary, the correct value of $x$ explains the training data, and accommodates the test data (so that plausible unseen data is allowed), but it does not spread the distribution needlessly wide over the total space of attribute combinations.

For the mushroom data set, the data is randomly divided into training and test sets, with 80% of the data assigned to the training set. Note the average probability of observed data decreases as $x$ increases (Fig. 7) since more of the mass is shifted onto unobserved data points.

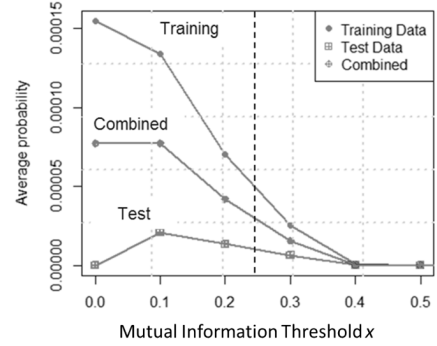

Fig. 7. Average probability of training data, test data, and all observed data combined, as a function of NMI threshold.

Maximizing $P(D|M(x))$, the quantity in equation (6), the optimal threshold for the mushroom data is found to be $x = 0.243$. The plot of $\log(P(D|M(x)))$ is shown in Fig. 8. For $x < 0.243$, $P(D|M(x)) = 0$ since at least some of the test data is not accounted for; this is the idea of the left side of Fig. 6. As $x$ increases past 0.243, $P(D|M(x))$ rapidly decreases monotonically. Note it is also possible to repeat the analysis for multiple partitionings of the data into test and training sets to improve robustness of the threshold determination.

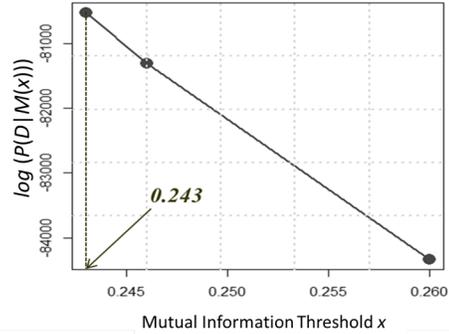

Fig. 8. Plot of $\log(P(D|M(x)))$ as a function of NMI threshold.

An interesting property of this solution is related to the entropy of the clique tree. Entropy is a function of the model $M(x)$, since the threshold determines clique tree structure. It can be calculated in terms of vertex and edge clique entropies:

$$H(M(x)) = \sum_{i \in V} H(C_i) - \sum_{ij \in E} H(C_{ij}) \quad (7)$$

where $C_i$ is a clique vertex in the clique tree and $C_{ij}$ is an edge clique (separator set). This expression reduces the calculation of entropy for a high dimensional distribution to the calculation of individual clique entropies which are simpler to calculate.

As expected, the entropy of the clique tree probability distribution increases as the threshold *x* increases (Fig. 9). In this plot the dashed line marks the location of the optimal NMI threshold where the feasible region for the threshold is to the right of this boundary. As this result indicates, the optimal threshold that maximizes the posterior probability corresponds to minimum entropy solution in the domain where the test data has non-zero probability.

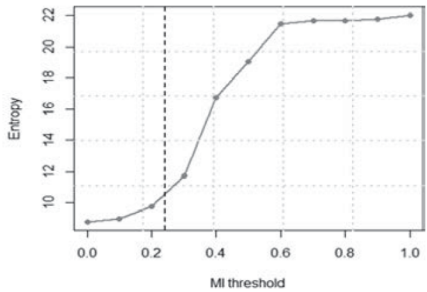

Fig. 9. Entropy of the clique tree probability distribution as a function of mutual information (MI) threshold. The dotted line shows the location of the optimal threshold 0.243.

Using the optimal threshold of 0.243, the corresponding optimal clique tree structure is shown in Fig. 10. Each node represents a clique, and is labeled with its attributes. The separator sets (the overlapping attributes between cliques) are not labeled on the graph; however, the cliques are linked if they have at least one attribute in common, and the separator sets are easily inferred by examining the node labels. Note that a few attributes are independent of all the others, but the typical clique size is in the range of 6-10 attributes.

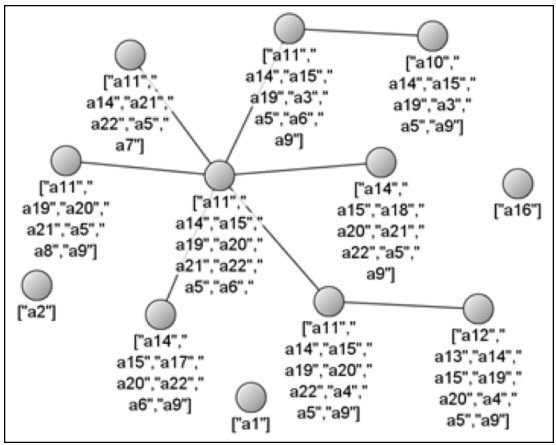

Fig. 10. Optimal clique tree structure for the mushroom data set.

## 3 Clustering and anomaly detection

Using the clique tree decomposition, we can examine data from the perspective of clique-based clustering, and discuss how cliques provide insight into anomaly detection and characterization.

### 3.1 Clique-based clustering

Clustering categorical data is challenging since many clustering approaches rely on distance metrics, and are therefore inapplicable to categorical data. For categorical data, some clustering approaches aim to find one optimal clustering ([10], [11], [12]). By contrast, the clique tree structure promotes alternative clusterings based on subspaces defined by the cliques of the strongly coupled attributes.

Clique-based clusterings provide a natural indexing mechanism for data where each clique provides an index and each data point is described by the particular combination of attributes that belong to that clique. For example, consider a clique $C_{14,17,6}$ formed by strongly coupled attributes (14,17,6) whose value multiplicities are (9,3,3). Based on the multiplicities of the variables involved there are 81 possible value combinations in the clique subspace. However, when the mushrooms are clustered according to their attribute combinations, only 11 clusters are found, with membership distribution shown in Fig. 11. Furthermore, the sizes of these clusters are highly skewed in that only 5 of the 11 clusters have a significant population.

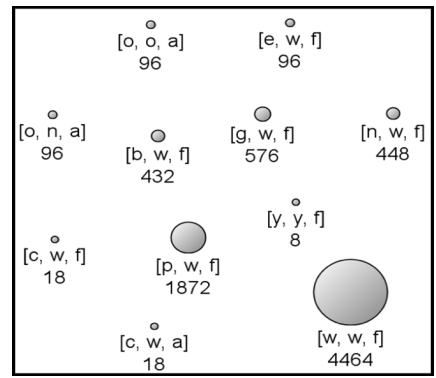

Fig. 11. Cluster memberships for a clique formed by attributes 14, 17, and 6. Each cluster is labeled by the ordered triplet of data values for the three attributes, as well as the membership count. Of the 81 possible attribute value combinations in this attribute subspace, only 11 value combinations were observed in the data.

For the clique structure in Fig 10, each mushroom would have potentially 12 indices defined by the 12 vertex cliques. The indices can be used to infer links between entities to form an association graph for data points. For example, a similarity measure between data points can be defined by measuring how often two data points fall into same clusters (share the same indices) according to different clique clusterings. The use of clique indexing for clustering categorical data is a promising direction for future exploration as it does not require a distance metric in categorical space.

## 3.2 Anomaly detection

One aspect of the "big data problem" is the challenge of finding events of interest, such as suspicious events, from very large data sets. Rule-based and classification-based approaches can assist this process by labeling some types of events, namely those with attribute combinations corresponding to a previously identified type of behavior. Anomaly detection is a more general approach to finding events of interest. In this approach, there is no ground truth to work from. In fact, part of the motivation for taking this approach is to discover anomalous or suspicious events which have not been previously characterized as such.

The intuition behind the anomaly detection approach is that most people or entities are engaged in innocent (or uninteresting) behavior most of the time. Put another way, suspicious events should be unusual events. A few remarks are necessary about this assertion: First, some suspicious or malicious events may actually be relatively common; however, we assume that commonplace events are detectable by other means, and it is not the goal of an anomaly detector to discover them. Second, unusual events may encompass many innocuous events as well as suspicious ones; a variety of one-time or idiosyncratic behaviors may be observed which are perfectly innocent. Even with these caveats, the important point is that suspicious events should be concentrated toward the low tail of the probability distribution.

The clique-based probability distribution derived in Section 2 assigns a probability value to all possible attribute value combinations, and thus can be used to find anomalous events. Further, the clique decomposition such as that in equation (4) also provides insight into why a data point is anomalous, by examining the probability values of each of the factors that comprise the right-hand side of the probability expression. This insight may assist in further investigation of the event. Returning to the example of the mushroom data and using Fig. 11, the distribution for clique $C_{14,17,6}$ shows that most of the mushrooms belong to 5 clusters, whereas 6 of the clusters have low membership. Attribute values corresponding to these clusters are possibly indicative of anomalous mushrooms.

In comparing clique probability values to elucidate the reason for an anomaly, it should be noted that cliques comprising more attributes will exhibit lower probability values in general. This is because there are more possible attribute value combinations and thus the data is distributed across a larger attribute space. Therefore, to perform clique comparisons, the clique probabilities should be normalized. One way to achieve this is to derive percentiles (or cumulative distribution functions) from the probability distributions for each clique.

Taking the investigative process one step further, when a clique is identified whose value combination was the reason for the anomaly, the analyst may query for other events with the same attribute value combination. This returns the cluster of similar events according to the ideas of Section 3.1. These other events may or may not be considered anomalies; their overall probabilities also depend on probability values in other cliques. Whether the similar events are anomalous or not, viewing the events as a group based on the clique clustering may be informative in characterizing behavior patterns.

## 4 Related work

Anomaly detection is one of the fundamental challenges in statistical inference with a rich literature. A recent survey of this field is provided in [13]. One of the difficulties in anomaly detection is the fact that training of an anomaly detector has to be performed without a reliance on ground truth as anomalies by definition do not form a cohesive class. Therefore, there is no single right answer for what constitutes an anomaly, and performance of anomaly detection is therefore harder to adjudicate than classification which can be trained from labeled classes. Approaches to anomaly detection in categorical attribute spaces must also overcome the lack of a distance metric. An alternative parameter free anomaly detection approach based on data compression is found in [14] and [15]. Work in these references relies on minimum description length and does not require a distance metric.

The method presented in this paper falls into the category of likelihood-based approaches. There are similarities between the approach presented and the method proposed in [16], in that probabilities are estimated using decomposition into lower dimensional spaces using groups of related attributes. There are also similarities to Bayesian network based anomaly detection proposed in [17] where structure of the network is determined by domain expert input. In comparison, the method we propose does not require any human input to specify the structure of the probability distribution. The main distinctions are that our approach optimizes the (often overlapping) subsets of attributes in the probability decomposition using a clique tree structure. Although we propose a method to automate the derivation of the optimal threshold and thus the probability distribution, the user may still choose to shift the threshold, or may combine the results of using different thresholds by using Bayesian model averaging. In this way, the user's domain knowledge is allowed to influence the probability estimations. Using attribute subspaces for analysis has some similarity to another approach [18] that is based on subspace based anomaly detection. A key difference of the presented approach is that it is a probabilistic model that can be used for any combination of categorical and numerical data.

## 5 Implementation

While computational aspects of the problem are not the main focus of this paper, the probability modeling, anomaly detection, and classification methods described in this paper were implemented using a scalable platform, Socrates [19]. All steps of the computations including computing the mutual information and clique tree optimization are parallelized in Socrates to accommodate large data sets.

## 6 Conclusions

In this paper, a clique tree approach to categorical data analysis has been presented with particular focus on the problem of learning of the clique tree structure for probability modeling and anomaly detection. The clique tree approach produces a probability model which exploits variable dependencies to decompose the joint probability into a product of joint probabilities in lower dimensions. By using lower dimensional subspaces, the probability model overcomes the problem of data sparsity, or the curse of dimensionality. At the same time, the probability decomposition provides clear anomaly signatures since it judges the likelihood of a data point by its conformance to expected variable dependencies in the lower dimensional subspaces. Finally, it has been shown that it is possible to use a Bayesian approach to determine a threshold that specifies the optimal structure of the clique tree. The optimal clique tree structure results in the probability model which best balances the competing requirements of fidelity to observed data and accommodating previously unseen data values.